**DRAFT VERSION**

# A Comparison of Human and ChatGPT Classification Performance on Complex Social Media Data

Authors: Breanna E. Green, Ashley L. Shea, Pengfei Zhao, Drew B. Margolin

## Abstract

Introduction. Generative artificial intelligence tools, like ChatGPT, are an increasingly utilized resource among computational social scientists. Nevertheless, there remains space for improved understanding of the performance of ChatGPT in complex tasks such as classifying and annotating datasets containing nuanced language.

Method. In this paper, we measure the performance of GPT-4 on one such task and compare results to human annotators. We investigate ChatGPT versions 3.5, 4, and 4o to examine performance given rapid changes in technological advancement of large language models. We employ a dataset containing human-annotated comments from YouTube and X. We craft four prompt styles as input and evaluate precision, recall, and F1 scores.

Analysis. Both quantitative and qualitative evaluations of results demonstrate that while including label definitions in prompts may help performance, overall GPT-4 has difficulty classifying nuanced language.

Results. Qualitative analysis reveals four specific findings: 1) cultural euphemisms are too nuanced for GPT-4 to understand, 2) interpreting the type of 'internet speak' found on social media platforms is a challenge, 3) GPT-4 falters in determining who or what is the target of directed attacks (e.g. the content or the user), and 4) the rationale GPT-4 provides is inconsistent in logic.

Conclusion. Our results suggest the use of ChatGPT in classification tasks involving nuanced language should be conducted with prudence.

# Introduction

Social media is rife with problematic content. This includes toxic speech such as harassment and hateful rhetoric, but also other forms of objectionable content that people find morally wrong to share, such as disinformation (Vahed et al., 2024). A common strategy for combating problematic content is the use of a discursive objection tactic, where individuals confront offensive comments through speech. These tactics differ in their approach to demarcating normative boundaries and often incorporate nuanced and important cultural references, whether implicitly or explicitly stated (Huang et al., 2023; Shea et al., 2024). Investigating and classifying such tactics at scale and across various social media platforms could be possible with advanced generative artificial intelligence (GenAI) tools. However, classifying such tactics involves complex, nuanced, and context-sensitive judgments and knowledge of important cultural euphemisms. This presents a unique opportunity and case study for comparing human and machine performance in handling complex social media data.

Are GenAI tools capable of comprehending the variety of problematic content that pervades social media? Research has demonstrated that advanced GenAI tools, like OpenAI's ChatGPT (Schulman et al., 2022) have the potential to perform classification and annotation tasks with accuracy comparable to or exceeding that of human annotators (Ding et al., 2023; Gilardi et al., 2023; Mellon et al., 2024; Mirza et al., 2024) and in a cost-effective manner (Wang et al., 2021). However, these large language models sometimes produce basic errors, particularly in understanding and rationalizing decisions (West et al., 2023). Therefore, we are particularly interested in how ChatGPT performs when comprehending and classifying the type of nuanced language used to combat problematic content online–a critical area of study to advance our understanding of AI performance in culturally embedded language, which is important for many research questions in computational social science.

This study is the first to our knowledge to explore the performance of ChatGPT in discerning and categorizing objection tactics and whether newer models are more effective than earlier versions in doing so. To fill this gap, we: 1) evaluate the performance of GPT-3.5, 4, and 4o on the classification of seven different objection tactics to determine which performed best, 2) discern the differences between the output of the best performer- GPT-4- with that of human annotators, and 3) conduct a qualitative assessment of the model's rationale for labeling each comment. Specifically, our study aims to answer the following research questions:
- RQ1: a) What is ChatGPT's performance in classifying objection tactics, and b) how does its performance vary with different prompts?
- RQ2: How do ChatGPT's annotations of objection strategies compare to human annotations (i.e., MTukers)?

# Related Work

## The Importance of Objection Tactics

Social media comment sections draw people from diverse backgrounds and with conflicting normative preferences together in shared spaces to discuss a variety of digital content. While echo chambers (Scheibenzuber et al., 2023) and information cocoons (Wan & Thompson, 2022) can unite individuals with shared viewpoints, research also shows that popular social media channels and trends can attract people with different ideologies who then inhabit shared online spaces (Wu & Resnick, 2021). When this happens, individuals will often encounter speech that they find objectionable and worthy of confrontation.

Collectively, objectionable and problematic content can undermine both individual well-being and the integrity of democratic processes (Rossini, 2022; Vitak et al., 2017). Although platforms employ strategies such as content removal and user bans to mitigate online toxicity, the use of top-down, platform-based moderation raises concerns about infringement on free speech and corporate accountability (Mathew et al., 2019). Additionally, differing interpretations of objectionable content, such as hate speech (Alkiviadou, 2019), combined with the vast volume of user-generated comments on social media, often make platform-based solutions inadequate. When individuals decide to engage in their own "expressive citizenship" (Gagrčin et al., 2022) they can employ a discursive objection tactic, defined as a communicative attempt to confront and stop what an individual perceives to be "wrong" or "bad" speech (Shea et al., 2024). This "stop command" signals to the original speaker and onlookers that this kind of communication should be changed or stopped.

## Challenges of Human Annotators Identifying Objection Tactics

In order to inform targeted educational and intervention programs for facilitating safer and healthy online discourses, it is necessary to understand and evaluate the real-world effectiveness of these tactics. However, manually classifying them is challenging. First, the process is time-consuming. Depending on the forum or context, discursive resistance to objectionable speech can be a rare phenomenon. Thus, to identify and classify even modest numbers of objections, human coders may need to read a large number of responses. A second problem is the mental and emotional strain for humans engaging with such content (Schöpke-Gonzalez et al., 2022). The conversations between offenders and objectors often contain offensive and harmful content, which can negatively affect the mental health of annotators. Moreover, the varying norms and cultures across online platforms and communities may result in different adoption and reception of these strategies (Waterloo et al., 2018). Variation also extends beyond platform-specific norms, and can exist at the individual level. For example, (Plank, 2022) argues that human label variation in annotation due to disagreement, subjectivity or ambiguity in answers has a large impact on all stages of the machine learning pipeline. Other times, variation in human labelling might stem from the annotator's demographics, like gender, race and age (Goyal et al., 2022; Larimore et al., 2021; Sap et al., 2022).

Annotating these tactics manually across diverse platforms, topics, contexts, and cultures demands extensive resources and a considerable workforce. These challenges underscore the need for scalable, accurate classification methods that can reduce human-related burdens. Being able to train a model that identifies distinct objection tactics at scale would be the first step in understanding the effects that such tactics have over time and across communities with different cultures and norms.

## Utilizing Generative AI & Large Language Models

Identifying objection tactics used to stop problematic content is challenging. Shea et al. (2024) show that human annotators require a level of comprehension and training that can often be costly. Common practice, before the increased use of GenAI, included training machine learning classifiers on gold standard labels (often through crowd-sourced labeling efforts) (Mathew et al., 2019; Shea et al., 2024). Traditional machine learning may be effective, but only after a large number of training examples have been identified, and even then there remains the risk that new conversational contexts limit generalizability. What is needed is a method that requires a modest number of examples for training that can be applied at scale over a large number of comments. This is a task for which large language models (LLMs) are well suited. There is preliminary evidence that suggests LLMs such as OpenAI's ChatGPT, Google Bard, or Anthropic's Claude can perform text labeling tasks at similar or better levels to that of human annotators, such as university students or paid Amazon Mechanical Turk workers (MTurkers) (Mellon et al., 2024; Mirza et al., 2024; Rescala et al., 2024). But is this true when the content requiring annotation is nuanced, and both contextually and culturally rich?

ChatGPT has been helping to reshape the way academic research can be conducted. Using natural language processing, ChatGPT is an LLM trained on an expansive amount of text data – enough to be able to simulate human understanding and generate human-like responses in return (Yin et al., 2024). Already, technology has advanced since the emergence of GPT-3 in 2020. Dale (2021) described the media attention that GPT-3 received as a mix of "awe and concern". The awe stemming from GPT-3 having a surprisingly coherent text-generation in some situations. The concern stemming from harmful unintended effects if and when this software with human-like intelligence is used by bad or even indifferent actors. However, studies focused on demonstrating the potential of ChatGPT tell us that despite its sophistication, its performance remains inconsistent . For example, (Reiss, 2023) used gpt-3.5-turbo to classify 234 German-based website text snippets as 'News' or 'not News'. Reiss used a combination of 10 different prompt instructions, 2 temperature parameters for the model (0.25 and 1), and 10 repetitions of classification task to determine if these differences affect consistency of the models' output. As a higher temperature is meant to induce more randomness in output, Reiss' findings reinforced that point with lower consistency at temperature of 1. He questioned if a newer model might be able to provide more consistent output, given the same, or similar, parameters.

A recent model of ChatGPT - GPT-4 - has been found to outperform other LLMs like Claude, Bard, and GPT-3.5 in reasoning, logic, facts, coding, language, and more (Borji & Mohammadian, 2023). Achiam et al. (2023) highlight the impact of technological advancement in LLMs. GPT-4 offers significant improvements over GPT-3.5 in terms of model sophistication and performance, making it a superior choice for classification tasks (Bubeck et al., 2023). The advanced architecture of GPT-4 should allow it to capture a wider range of linguistic nuances and process complex language patterns more effectively. This capability is crucial for accurately classifying text where subtle differences in wording or context can greatly impact the outcome. Additionally, GPT-4's deeper neural network can potentially enhance its ability to generalize across various data types, leading to more reliable predictions, especially when dealing with novel or niche examples.

However, as Ollion et al. (2023) caution, we should still mind the hype around using ChatGPT for classification tasks. Specifically, they carried out a systematic review of studies that applied GPT-4 for zero-shot or few-shot text classification across various fields, including Psychology, Political Science, Computation Social Science, and Communication. While their review bolstered the idea that these LLMs perform relatively well compared to humans and often outperformed preLLM automatic annotation methods, they also highlight a number of concerns. These include (1) LLMs were not necessarily the best models when compared to finely-tuned transformer models, (2) the range of performance scores is quite wide across papers which calls for thorough validation, (3) GPT-4 tended to identify more false positives than false negatives, producing a higher recall than precision, and lastly (4) the evaluation metrics across papers also varied widely, making comparison difficult.

Now with OpenAI's GPT-4o, currently the most recent version which began rollout around May 13, 2024 we can evaluate if the concerns highlighted by Ollion et al. (2023) continue to a more advanced model. Building on the capabilities of GPT4, GPT-4o further enhances performance in classification tasks through advanced alignment and optimization techniques. GPT-4o incorporates more sophisticated algorithms and a larger training dataset, enabling it to handle a broader range of linguistic variations and contextual subtleties. It can process and interpret longer text sequences, offering deeper contextual understanding and memory capacity. With its superior ability to interpret user intent and manage ambiguous inputs, ChatGPT-4o has the potential to provide unmatched performance in classification tasks, even surpassing both GPT-3.5 and GPT-4 in accuracy, efficiency, and versatility.

# Dataset & Methods

In the first phase of this study, we assess the efficacy of all three versions of GPT - GPT-3.5, 4, and 4o - in labeling objection tactics (Shea et al., 2024). To do this, we crafted four prompts as input and used each prompt with all three ChatGPT models (GPT-3.5, 4, and 4o) to classify 560 objection tactics. We then took the results of this comparison across models to focus on the performance of the best performing model, GPT-4, for further insights. In this second phase of the study, we compared GPT-4 results to a measure of agreement between MTurkers. Then, we conducted a qualitative analysis of the GPT-4 results, specifically. Data were drawn from both YouTube and Twitter in 2021, prior to Twitter's ownership change 1 and the platform renamed to X (herein, we will refer to Twitter as X).

Selected GPT Model Parameters. Of the currently available ChatGPT models, we utilized GPT 3.5, 4, and 4o. Across each prompt experiment and GPT version, we maintained consistent parameter choices: all default settings except for temperature set at 0.2 (default is set at 1.0). The temperature parameter ranges from 0 to 2, where larger values will make the output more random. Lower values like 0.2 will make the output more focused and deterministic.

## Prompt Types

Prompt engineering is the iterative process of crafting and clarifying input – prompts – to garner a desired response from an LLM. To address our research questions, we developed four prompts, paying special attention to the ChatGPT prompting framework outlined in White et al. (2023).

| Prompt | Design Details | Example Format |
|---|---|---|
| Prompt 1 | Provide all strategy labels only and an exemplar comment. Return an assigned strategy. | Here are 7 distinct objection strategies that users employ in online comment sections: LABELS. Here are quoted examples of comments using each strategy with the strategy noted in parentheses following each example. The end of the example is denoted with a semicolon: EXEMPLARS (LABELS). Based on these examples, which of the 7 strategies would you assign to the following comment? Only return the strategy name you assign: NEW COMMENT. |
| Prompt 2 | Provide all strategy definitions only and an exemplar comment. Return an assigned strategy. | You are a researcher labeling data. There are 7 distinct objection strategies that users employ in online comment sections. Below are the definitions of each strategy in square brackets followed by examples of comments using each strategy in parentheses. The end of the example is denoted with a semicolon: [DEFINITIONS] EXEMPLARS. Based on these examples, which of the 7 strategies would you assign to the following comment? Only return the strategy name you assign: NEW COMMENT. |
| Prompt 3 | Provide all strategy labels only and an exemplar comment. Allow for "No Match" response. | Here are 7 distinct objection strategies that users employ in online comment sections: LABELS. Here are quoted examples of comments using each strategy with the strategy noted in parentheses following each example. The end of the example is denoted with a semicolon: EXEMPLARS (LABELS). Based on these examples, which of the 7 strategies would you assign to the following comment? If the comment does not align with any of the strategies, return "No Match". Only return the label you assign: NEW COMMENT. |
| Prompt 4 | Individual labels: Check all comments for each strategy based on definition. Return "Yes" or "No", and reasoning. | You are a researcher labeling comments as either INDIVIDUAL LABEL or not. For a comment to be labeled as INDIVIDUAL LABEL, you are looking for INDIVIDUAL DEFINITION. Based on this information, would you label this comment as INDIVIDUAL LABEL? Please return either a "Yes" or "No" then a semicolon then your reasoning: NEW COMMENT. |

Table 1: Outlines the four (4) prompt designs used in the study. For each prompt design, we demonstrate the details used for crating the prompt, as well as the format of the prompt input for GPT-4.

Prompts 1, 2 and 3 are derivations of each other, built to examine which element (if any) of the prompt might induce the most impact on the results: tactic labels or definitions, and the option to return "No Match". We held constant providing each of the GPT versions an exemplar comment representative of the tactic in question. Prompt 1 provided only strategy labels and exemplars, without offering tactic definitions to GPT-4. On objections tactics that received poor F1 scores across the board - moral corruption and content threat - this prompt type performed somewhat better. Under this prompt type, for every tactic except for logical disqualification, we find the precision to be greater than recall. Prompt 3 follows a similar form to that of prompt 1, which makes sense given both do not provide definitions for tactics.

Unlike the first three prompts, prompt 4 requires GPT-4 to evaluate each of the seven tactics using a binary classification task (e.g. is the comment moral corruption, 'Yes' or 'No'?), as well as provide reasoning or explanations for the classification. The format of prompt 4 follows closest to the instructions presented to MTurkers in Shea et al. (2024). Each MTurker was trained on specifically one of the seven tactics, and asked to determine if that tactic was present in a comment (e.g. answer "Yes" or "No").

Measures of Performance. Given the imbalance of classes among our objection tactics dataset and for brevity, we report F-1 scores when comparing the three model versions. We then report precision, recall, and F1 scores when comparing across objection tactics and prompt types for one model version. The previously established human annotations were used as gold standard or ground-truth labels to compare against. F1 scores between 0.5 and 0.8 are considered average.

As a measure of inter-rater reliability between MTurkers for the objection tactic dataset, we used Krippendorf's alpha of agreement $K\alpha$ (Krippendorff, 1970) is used as a comparison measure against the F1 scores. We note that $K\alpha$ scores from 0.67 to 0.79 are considered average. We compare $K\alpha$ and F1 scores to evaluate performance.

# Results

## Examining the Impact of Technological Advancement of GenAI and LLMs

As expected, the progression from GPT-3.5 to GPT-4 led to noticeable improvements in classification accuracy, with GPT-4 demonstrating a better ability to handle complex language patterns and contextual subtleties. These enhancements are consistent with the advancements in the model's architecture, which allows it to capture more intricate details and nuances within the data. However, the results also indicate that while model sophistication can lead to better outcomes, the relationship between model complexity and performance is not linear and may plateau or even regress in certain scenarios. For example, contrary to initial expectations, GPT-4o did not outperform its predecessors as significantly as anticipated, suggesting that the improvements in sophistication do not always directly translate to superior performance in classification.

| Objection Tactic | GPT Version | Prompt F1 Scores | | | |
|---|---|---|---|---|---|
| | | 1 | 2 | 3 | 4 |
| Moral corruption | GPT 3.5 | 0.370 | 0.500 | 0.364 | 0.249 |
| | GPT-4 | 0.391 | 0.480 | 0.304 | 0.372 |
| | GPT-4o | 0.487 | 0.364 | 0.480 | - |
| Logical disqualification | GPT 3.5 | 0.675 | 0.592 | 0.411 | 0.357 |
| | GPT-4 | 0.682 | 0.619 | 0.647 | 0.453 |
| | GPT-4o | 0.648 | 0.644 | 0.665 | 0.014 |
| Physical threat | GPT 3.5 | 0.765 | 0.714 | 0.824 | 0.619 |
| | GPT-4 | 0.786 | 0.833 | 0.741 | 0.769 |
| | GPT-4o | 0.909 | 0.895 | 0.857 | - |
| Ad hominem | GPT 3.5 | 0.727 | 0.377 | 0.723 | 0.693 |
| | GPT-4 | 0.696 | 0.551 | 0.706 | 0.741 |
| | GPT-4o | 0.621 | 0.645 | 0.680 | 0.459 |
| Content threat | GPT 3.5 | 0.149 | 0.385 | 0.242 | 0.262 |
| | GPT-4 | 0.409 | 0.143 | 0.325 | 0.279 |
| | GPT-4o | 0.344 | 0.273 | 0.312 | 0.061 |
| Self control | GPT 3.5 | 0.554 | 0.566 | 0.481 | 0.540 |
| | GPT-4 | 0.667 | 0.773 | 0.450 | 0.760 |
| | GPT-4o | 0.697 | 0.863 | 0.720 | 0.071 |
| Space control | GPT 3.5 | 0.213 | 0.567 | 0.275 | 0.284 |
| | GPT-4 | 0.649 | 0.737 | 0.667 | 0.634 |
| | GPT-4o | 0.613 | 0.729 | 0.618 | 0.600 |

Table 2: Displays the F1 scores for each GPT version and prompt for comparison.

Across prompts 1, 2 and 3, GPT-4o was able to produce comparable or better F1 scores than its predecessors for each of the objection tactics. Prompt 4 seemed to have been the most difficult prompt version for GPT-4o. For both moral corruption and physical threat GPT-4o performed abysmally and was unable to identify any of the comments with the correct class. For logical disqualification, content threat, and self control, GPT-4o also did quite poorly – especially in relation to its predecessors GPT 3.5 and 4. Space control resulted in a comparable F1 score to that of GPT-4, yet did not surpass it.

One possible explanation for the under performance of GPT-4o relative to expectations is the increased complexity and fine-tuning of the model may have inadvertently introduced new challenges or biases. As models become more intricate, they may also become more sensitive to the specific characteristics of the training data, leading to over-fitting or a decrease in generalizability. Additionally, the fine-tuning process, while intended to improve performance, could have made the model more specialized in ways that are not entirely aligned with the requirements of the nuanced classification task at hand. This specialization may have limited the model's ability to adapt to the specific nuances and variations present in the objection tactic data.

Another consideration involves the task itself may not fully benefit from the additional capabilities offered by GPT-4o. While the newest version is designed to handle a broader range of linguistic subtleties and more complex inputs, the specific nuances of the classification task might have been better suited to the capabilities of GPT-4, for reasons that we as researchers may not be able to directly manipulate. In this case, the improvements in GPT-4o could be either unnecessary for the task or even detrimental if they lead to the model focusing on irrelevant aspects of the data. This insight underscores the importance of task-specific evaluation when deploying increasingly sophisticated AI models and suggests that more advanced does not always equate to better performance in every context.

## Assessing Differences Across Prompts and Objections Tactics Using GPT-4

Table 3 presents the measures of performance across each of the four prompt types and seven objection tactics, based on 560 comments. Our findings show that no single objection tactic or prompt type produces an average desirable F1 score > 0.8 across the board. Therefore, we examine results first by prompt and second by objection tactic (see Figure 1).

Prompts. Prompt 2, which provides only the tactic definition and an exemplar, generally performs well compared to the other prompt types - achieving the highest F1 scores in 3/7 of the objection tactics. It is the only prompt type that achieved an F1 score greater than 0.8 for any of the tactics (i.e. physical threat F1 = 0.833). However, in the case of content threat and ad hominem, it performed quite poorly with F1 scores of 0.551 and 0.143, respectively. In both cases, precision was much greater than recall, implying that we have many false negatives. This was particularly prominent in the case of content threat (recall = 0.080; precision = 0.636). Excluding content threat and space control, prompt 4 demonstrates more balanced precision and recall scores for each tactic. However, in this case 4/7 tactics have a higher recall than precision, indicating concern for a greater number of false positives.

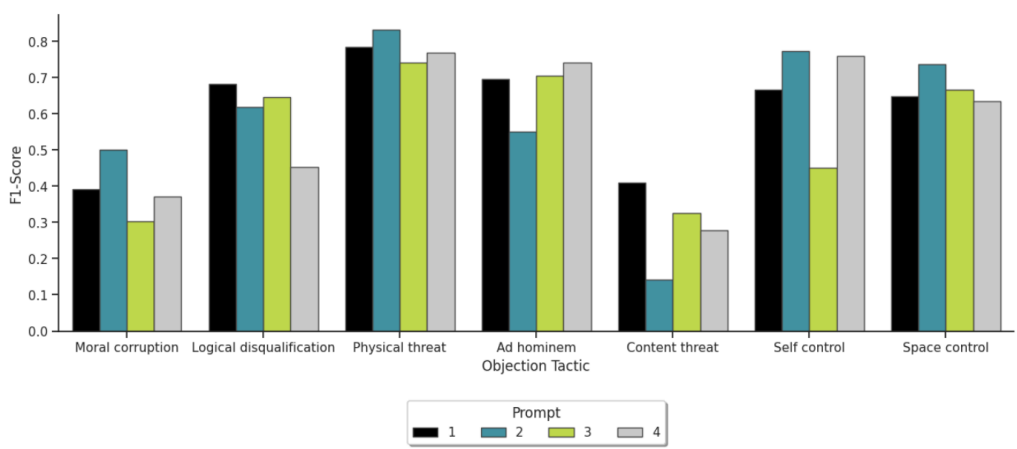

Figure 1: Displays the F1-scores of each prompt type by objection tactic.

|  |  | MC (n=60) | LD (n=146) | PT (n=17) | AH (n=270) | CT (n=87) | SfC (n=27) | SpC (n=40) |
|---|---|---|---|---|---|---|---|---|
| Prompt 1 | Precision | 0.630 | 0.575 | **1.000** | 0.795⋆ | 0.600 | 0.762 | 0.706 |
|  | Recall | 0.283 | **0.836** | 0.647⋆ | 0.619 | 0.310 | 0.593 | 0.600 |
|  | F1 Score | 0.391 | 0.682 | **0.786** | 0.696⋆ | 0.409 | 0.667 | 0.649 |
| Prompt 2 | Precision | 0.591 | 0.459 | 0.789 | 0.853⋆ | 0.636 | **1.000** | 0.778 |
|  | Recall | 0.433 | **0.952** | 0.882⋆ | 0.407 | 0.080 | 0.630 | 0.700 |
|  | F1 Score | 0.500 | 0.619 | **0.833** | 0.551 | 0.143 | 0.773⋆ | 0.737 |
| Prompt 3 | Precision | 0.632 | 0.578 | **1.000** | 0.774⋆ | 0.633 | 0.692 | 0.750 |
|  | Recall | 0.200 | **0.733** | 0.588 | 0.648⋆ | 0.218 | 0.333 | 0.600 |
|  | F1 Score | 0.304 | 0.647 | **0.741** | 0.706⋆ | 0.325 | 0.450 | 0.667 |
| Prompt 4 | Precision | 0.396 | 0.484 | 0.682 | 0.696⋆ | 0.172 | **0.826** | 0.470 |
|  | Recall | 0.350 | 0.425 | 0.882⋆ | 0.796 | 0.736 | 0.704 | **0.975** |
|  | F1 Score | 0.372 | 0.453 | **0.769** | 0.741 | 0.279 | 0.760⋆ | 0.634 |

Table 3: The precision, recall and F1 score for each prompt and tactic: moral corruption (MC), logical disqualification (LD), physical threat (PT), ad hominem (AD), content threat (CT), self control (SfC) and space control (SpC). The largest values per row are bolded, while the second largest values are followed by a star symbol (⋆).

Objections Tactics. Results between objection tactics were more prominent than across prompt types. Physical threat – with a small overall subset size (n=17) – consistently has the highest F1 score across prompts, with little fluctuation. Space control also demonstrates consistency across prompts, maintaining F1 scores between 0.634 and 0.737. Content threat and moral corruption have fluctuating F1 scores based on prompt type but are overall consistently lower, relative to the other objection tactics. Logical disqualification results in the highest recall scores among tactics and across prompts, except in the case of prompt 4. This would imply that in attempting to classify logical disqualification, many of the ground-truth comments have been accurately detected but we also have many false positives.

GPT-4 & Human Agreement. Results comparing the inter-rater reliability of MTurkers (Kα), can be found in Table 4, along with the precision, recall and F1 scores of prompt 4.

| Objection Tactic | MTurkers | Prompt 4 | | |
|---|---|---|---|---|
| | $K_\alpha$ | P | R | F1 |
| Space Control | 0.78 | 0.470 | 0.975 | 0.634 |
| Content threat | 0.76 | 0.172 | 0.736 | 0.279 |
| Self Control | 0.72 | 0.826 | 0.704 | 0.760 |
| Physical threat | 0.71 | 0.682 | 0.882 | 0.769 |
| Ad hominem | 0.65 | 0.696 | 0.796 | 0.741 |
| Logical disqualification | 0.63 | 0.484 | 0.425 | 0.453 |
| Moral corruption | 0.56 | 0.396 | 0.350 | 0.372 |

Table 4: Displays the Krippendorf's alpha ($K_\alpha$) from Shea et al. (2024) and prompt 4 precision, recall, and F1 score for comparison. Alpha represents a measure of the agreement between annotators for those who did pass the quiz. Table values sorted descending by alpha.

The most stark contrast between Kα and F1 scores is in content threat. Content threat had the highest agreement among humans (Kα = 0.78), but GPT-4 did poorly at identifying this tactic. Moreover, this poor performance on content threat was consistent for other prompts as well. Incorrect labeling is reflected in the higher recall and lower precision we see (0.736; 0.172). This indicates GPT-4 identified most of the true positive content threat comments but was also prone to identifying false positives or mistaking other types of comments as such. Inspection reveals that the problem is due to incorrectly characterizing many ad hominem attacks as content threat and vice versa (a point we will develop further in the next section). This difficulty in distinguishing these two can be seen in the results of prompt 2 as well. Space control was another tactic where humans had the highest agreement in coding (Kα = 0.78), and GPT-4 did not perform as well, though with not as stark of a difference. Here it is also a case of high recall but poor precision (0.975; 0.470). GPT-4 identifies space control where humans do not.

In cases where they both did well – self control and physical threat – there is a trade-off between precision and recall. However, both F1 scores are higher than Kα. There were no cases where GPT-4 performed markedly better than humans, but we note that it was quite strong in ad hominem with an F1 score of 0.741. While humans still did okay here, they did not do as well as other tactics and fell below an average range of Kα (< 0.67). Finally, moral corruption and logical disqualification were at the bottom for both humans and GPT-4.

## Understanding Discrepancies between GPT-4 Output and Internal Ground Truth

To understand why GPT-4 was better at catching and labeling certain tactics over others, our team conducted a thematic analysis of GPT reasoning output to better understand

**DRAFT VERSION**

discrepancies between human-annotated ground truth and the GPT-4 prescribed labels. Below, we report thematic insights that emerged.

Finding 1: Cultural euphemisms are too nuanced for GPT-4 to understand. GPT-4 lacks nuanced understanding of cultural references, including "coded" insults which are important when classifying ad hominem tactics. For instance, the phrase "LET'S GO BRANDON" is a cryptic euphemism for a much more vulgar insult directed at U.S. President Joe Biden and by extension, anyone who supports him.

Finding 1: Cultural Euphemsims
'@user LETS GO BRANDON!! LETS GO!!'
'@User walks into a bar -your Dunnig Krueger is showing. So now you essentially say that the kkk didn't originate from the democrats? You're the one who mentioned them in the first place and now you say I'm lying about them? Who's the liar, -snowflake. I haven't said a word about your imaginary paper or friend; but you call me a liar anyway. I quoted the en car t a dictionary definition of fascism and you again, don't like it and call me a liar. Typical. When your hand is called, it's always the other guy's fault; right?'
Table 5: Depicts examples of nuanced cultural euphemisms.

This phrase of attack is often used to dismiss people based on their presumed identity or political ideology (Long, 2021). When the phrase was observed in comments included in our sample (see Table 5), human annotators labeled such comments as an ad hominem attack in recognition of its derogatory aim. GPT, however, missed this labeling, and when prompted for rationale, responded by saying that the comment "seems to be cheering on or supporting someone named Brandon." It is clear in this case, that cultural euphemisms- especially relatively new ones like this one which emerged in prominence in 2021- can pose challenges for automated models.

"Snowflake" is another cultural euphemism often used to attack another person deemed hypersensitive. It is typically used by conservatives when attacking perceived liberals and members of Generation Y to highlight a problematic intergenerational divide between people (Alyeksyeyeva, 2017). As Table 5 shows, the comment was labeled by human annotators as an ad hominem attack, but a content threat by GPT. Accusatory labels such as "Dunning Kruger", "liar", and "snowflake" were used to attack the person commenting rather than the content of the prior comment. However, GPT-4 cites the use of the word "liar" to disparage the comment. This mislabeling leads us to our second finding.

Finding 2: Interpreting Internet speak is difficult for GPT-4. Many online comments contain acronyms and emojis that convey specific meanings, but GPT-4 misses these. For example, the first comment shown in Table 6 uses an acronym ("STFU") to tell another person to stop talking and leave the space. Human annotators recognized this acronym stood for "shut the f - - - up" and classified the strategy as a space control objection tactic. However, GPT-4 labeled it a content threat, seemingly focusing on "red flag" words like murder, crime, and racist to classify the comment even though they were not used as expected within a content threat.

**DRAFT VERSION**

Finding 2: Acronyms & Emojis
'@user Oh stfu and go away no one cares any more. I care more about the murder and crime rates then being called racist.'
'@user bro stop drinking so much koolaid and stop watching skynews'
Table 6: Depicts example tweets containing either acroynms or emojis that GPT-4 could not assess in classification.

In the second example, a comment containing several clown and sheep emojis was annotated by human coders as an ad hominem attack because the icons were interpreted by the annotators as hurled insults akin to typing "clown" and "sheep". GPT-4 did not catch that these were ad hominem attacks. Instead, the comment was labeled a content threat, which takes us to our next finding.

Finding 3: GPT-4 has issues determining what (or who) is the target. When insulting labels are directed at another user, our classification typology states that it is an example of an ad hominem attack because the character or reputation of the commenter is being attacked instead of the content of a comment. However, when GPT-4 assessed known ad hominem attacks, it often deemed them content threats. For example, the first comment in Table 7 should have been labeled by GPT-4 as an ad hominem attack.

Finding 3: Target of attacks
'@user You're a Trump supporter. How much more dimwitted can you get?'
'@user Human Russian spies know America better than you traitor Republicans, is a fact.'
Table 7: Depicts example tweets wherein the user being replied to is the target of insult, rather than the content of the message.

We do not have data for what comment to which an individual was replying. The comment used an accusatory label ("dimwitted") to attack and dismiss the user based on their political affiliation ("Trump supporter"). However, GPT-4 classified it as a content threat, which should be applied only when the content of a comment or argument is directly attacked. Had the comment instead stated "the claim in your comment is dimwitted", then the accusatory label would indeed be directed at the content of the comment and the label of content threat would be appropriate. The second comment in Table 7 is another example where GPT-4 falsely labeled an ad hominem attack as a content threat. The use of the phrase "you traitor Republicans" is a clear attack on the person rather than the content. Our sample reveals that GPT-4 seems to struggle to determine the target of the objection (i.e., person vs. content), which is an essential difference that distinguishes ad hominem attacks and content threats.

Finding 4: GPT-4 is inconsistent in rationalizing decisions. When reviewing the rationale produced by GPT-4 following its classification, we noticed inconsistencies. For example, the top comment in Table 8 should be labeled a physical threat.

Finding 4: Contradiction in rationale
Physical threat

'@user Thomas Jefferson predicted this and said we'd need regular revolution. I'm ready.'
GPT-4 rationale
'While the comment implies a readiness for drastic action, it does not directly threaten or imply violence towards a specific person or their in-group.'
Table 8: Depicts example tweet containing sensationalized language that may be misinterpreted as physical harm or content threat by implying a need and readiness for civil war.

The justification made by GPT-4 in not labeling this comment as a physical threat, found below, included a clear contradiction. Despite stating that the comment implies readiness for drastic action, it then contradicts the statement by stating it does not imply violence.

# Discussion

## General Overview of Results.

Our motivation for conducting this study centered on evaluating the performance and efficiency of using AI tools – in this case ChatGPT – in a specific, niche annotation task. Specifically, we evaluated whether GPT-3.5, 4, and 4o were capable of classifying seven different objection tactics used in response to problematic content. We experimented with four different prompting types for the classification task and measured precision, recall, and F1 scores. For the fourth prompt, we incorporated a binary classification task per objection tactic similar to that conducted in Shea et al. (2024). We found that GPT-4 had markedly consistent improved performance over GPT3.5, but surprisingly GPT-4o did not. Moreover, when encountering a binary classification task (prompt 4), GPT-4o performed well below our initial expectations. In two of seven objection tactics (moral corruption and physical threat), the newest model was unable to discern any of the comments as the correct label. Conversely, its predecessor GPT-4 was able to complete the task. Thus, upon further examining the capabilities of GPT-4 specifically, we found that it provides consistent responses for particular objection tactics as opposed to inconsistent responses for others.

## Overview of Results for GPT-4.

Prompt 2 performed the best. It consisted of the strategy definition and an exemplar comment as input to GPT-4. The model was able to perform on par with human annotators for both the self control and physical threat objection tactics despite each tactic representing only a small percentage of the data. Moreover, regardless of the prompt type, GPT-4 had the best performance on physical threat. Prompt 3, which provided strategy labels and exemplars but no definition, performed noticeably worse than the other prompts.

Qualitative analysis reveals four significant themes of interest: 1) cultural euphemisms are too nuanced for GPT-4 to understand, 2) interpreting the language found on social media platforms ("internet speak") is also a challenge, 3) GPT-4 has issues determining who or what is the target of directed attacks (e.g. the content or the user), and 4) the rationale GPT-4 provides has inconsistencies in logic.

**DRAFT VERSION**

  The ability of LLMs like GPT to generate meaningful responses is inherently limited by the information they have access to. These models rely exclusively on patterns from their training data, meaning they lack awareness of knowledge or context not explicitly present in that data. As a result, their responses can miss crucial layers of meaning that humans derive from lived experience and implicit understanding. This limitation becomes especially apparent when models attempt to address topics requiring nuanced cultural knowledge, where much of the meaning is implied rather than directly conveyed through text.

  Moreover, language models struggle to interpret non-verbal cues, such as emojis and images, which carry emotional weight and context-sensitive meanings for humans. While these elements are often ambiguous, humans draw on cultural background and personal experience to assign emotional significance to them — something GPT cannot replicate. Even if certain patterns associated with these cues appear frequently in training data, the model's understanding remains superficial, as it processes them solely as tokens without capturing the emotional resonance they hold in the real world. Additionally, human responses are deeply influenced by context, such as situational dynamics, personal relationships, and emotional states, which exist beyond the scope of language alone. This makes it challenging for GPT to fully align with human thought and behavior, as it is limited to analyzing text without grasping the surrounding circumstances that inform meaning. Thus, there is a fundamental gap between the situational awareness needed for humanlike interaction and the purely linguistic nature of language models.

  This work also demonstrates how designing prompts for LLMs presents unique challenges, as the prompts effective for human use may not yield optimal results for LLMs. Humans naturally interpret language with nuance, relying on connotations, context, and implied meaning. However, LLMs process language literally, adhering to the strict definitions and standardized patterns found in their training data. This creates a gap between human and LLM interpretations, where ambiguous or imprecise prompts can lead to miscommunication or incorrect responses from the model. LLMs require precise, formal prompts because they lack the capacity to infer meaning beyond the words provided. While human coders can adjust their understanding to account for nuanced usage — ignoring denotations that disrupt meaning — LLMs are constrained by standard linguistic rules and definitions. For example, if a word like "accusatory" is used inaccurately in a prompt, the LLM may misinterpret it because it cannot bypass incorrect usage the way humans can. This suggests that reinforcement learning during prompt engineering may help improve the model's ability to handle imprecise language more flexibly. However, there is a trade-off between technical precision and accessibility when designing prompts. More sophisticated, technical language (e.g., "defamatory") often helps LLMs perform better, but such language may alienate human annotators who are not familiar with the meaning. While LLMs excel with dictionary-defined or formal vocabulary, they struggle with informal or context-dependent language, such as slang, which humans understand intuitively.

  Lastly, LLMs like GPT can generate responses that appear rational, but these are better understood as post-hoc rationalizations rather than true cognitive reasoning. While the models predict plausible outputs based on patterns in their training data, they lack the ability to intentionally reason or understand the meaning behind their responses. This creates an illusion of reasoning that raises important questions about the limits of AI-generated explanations and the transparency of these systems. Future research can explore how to improve the clarity of LLM processes, better align user expectations with what these models can and cannot explain, and investigate the ethical implications of relying on systems that simulate reasoning without genuine understanding.

## Implications.

The black-box nature of LLMs continues to be a concern when applied in computational social science settings (Thapa et al., 2023). While trained on vast amounts of general language from across the web, they may lack the domain-specific knowledge needed to annotate nuanced social media comments. Despite the findings for GPT-4o, it is surely possible that as newer models are deployed to the public, ChatGPT will get better at this task over time. However, in the case where researchers might need large-scale annotation support now and are looking to GenAI to fill this space, we demonstrate that this must be approached with caution.

For example, GPT-4 often conflated ad hominem and content threat, which are two tactics that are similar in form but require an understanding of relational intent. Ad hominems are replies that use accusatory labels to attack or smear the reputation of the user/person to whom one is replying. Content threats use accusatory labels to attack only the content, not the person. This relational intent, attacking a person or their comment, is precise, and therefore classifying it requires acute attention.

## Limitations.

While we establish that GPT-4 is less effective in classifying more nuanced language that can be found on social media platforms, a limitation of this study is that examining objection tactics alone might tell us little about the use of ChatGPT in other niche contexts. There might be social phenomena both nuanced in form and easily classifiable by ChatGPT. However, that is outside of the scope of the current work.

Furthermore, our study only examines four prompt types, wherein three of them are close in form. It is possible that we would observe different performances with a number of starkly different prompt types. Extending the diversity of prompt types through prompt engineering to further test the text features that contribute to increased performance measures of GPT-4 is essential (Kocoń et al., 2023). More advanced prompting methods not utilized in the current study, such as chain-ofthought prompting (Wei et al., 2022), could also be implemented in future work. Chain-of-though prompting, or querying the LLM for a series of intermediate reasoning steps, appears to greatly improve performance on tasks such as on arithmetic reasoning, commonsense, and symbolic reasoning tasks. Thus, it is possible that evoking a chain-of-thought prompting schema could help the model reason through niche annotation.

Lastly, we also recognize and acknowledge the overall size of the original objection tactic dataset is smaller than many of the common datasets used for NLP classification bench-marking. However, we believe the sparseness of the dataset further illuminates the need for automated large-scale detection and valid annotation efforts, as opposed to the many hours of manual labor taken on by the human researchers who generated the dataset.

# Conclusion

Problematic content on social media remains an issue worth tackling. The rise of GenAI across the research landscape has the potential to aid in the large-scale detection of this type of content. However, there is still room for improvement in the human-AI collaborative process. Especially in evaluating content that is much more nuanced and/or surreptitious. The current study adds to the growing body of literature exploring this relationship.

We found through multiple prompt testing that ChatGPT has a difficult time parsing out seven distinct objection tactics. Using distinct discursive objection tactics as a mechanism to explore our research questions, we found that GPT-4 performs well when a comment lacks references to cultural phenomena or normative expectations and moral rationale. For example, comments with clear directives and statements are correctly classified. However, comments on social media are rarely free from the social context in which they are influenced and influence. Thus, when it comes to relying on trained models to make sense of behavioral interactions rich with meaning-making, it still seems based on our findings that some degree of human-in-the-loop involvement is needed.

The rapid advancement of GenAI, including models like ChatGPT, has sparked a range of concerns across fields and industries. These concerns touch on ethical, social, economic, and technical dimensions, reflecting the profound impact that these technologies are beginning to have on society. As GenAI continues to evolve, these issues will require careful consideration and proactive management to ensure that the benefits of the technology are realized while mitigating potential harms. Echoing the sentiments of Dale (2021), we see that LLMs, and ChatGPT models specifically, are not 'devoid of practical application; far from it. But it means that some use cases are appropriate and some are not.' This work contributes to the ever-evolving literature around the appropriate use cases of ChatGPT – not only in the interests of computational social scientist but of various disciplines and for numerous applications.

Future work aims to better understand the 'reasoning' that GPT-4 gives in its classification schema. One avenue is to employ human annotators' to supplement their labeling of objection tactics with their reasoning, and then compare human reasoning to GPT-4 reasoning. Second, while fine-tuning GPT-4 is possible, many social science and humanities researchers interested in this tool may have difficulties engaging with the fine-tuning process. As we and other research (Chen et al., 2023; Marvin et al., 2023; Polak & Morgan, 2024) have shown, prompt engineering has a noticeable impact on the precision and recall of GPT-4 as a form of classification model. Lastly, we imagine the use of GPT-4 can support a feedback loop of improved definitions for fuzzy definitions of topics. Specifically, in this case, the results may allow researchers to better refine definitions of objection tactics without losing acknowledgment of nuance. This process, however, can be implemented in various research pipelines seeking to bolster the consistency of their label definitions.

# References

Achiam, J., Adler, S., Agarwal, S., Ahmad, L., Akkaya, I., Aleman, F. L., Almeida, D., Altenschmidt, J., Altman, S., Anadkat, S., et al. (2023). Gpt-4 technical report. arXiv preprint arXiv:2303.08774.

Alkiviadou, N. (2019). Hate speech on social media networks: Towards a regulatory framework? Information & Communications Technology Law, 28(1), 19–35.

Alyeksyeyeva, I. (2017). Defining snowflake in british post-brexit and us post-election public discourse. Science and Education a New Dimension, 39(143), 7–10.

Borji, A., & Mohammadian, M. (2023). Battle of the wordsmiths: Comparing chatgpt, gpt-4, claude, and bard. SSRN Electronic Journal.

Bubeck, S., Chandrasekaran, V., Eldan, R., Gehrke, J., Horvitz, E., Kamar, E., Lee, P., Lee, Y. T., Li, Y., Lundberg, S., et al. (2023). Sparks of artificial general intelligence: Early experiments with gpt-4. arXiv preprint arXiv:2303.12712.

Chen, B., Zhang, Z., Langrené, N., & Zhu, S. (2023). Unleashing the potential of prompt engineering in large language models: A comprehensive review. arXiv preprint arXiv:2310.14735.

Dale, R. (2021). Gpt-3: What's it good for? Natural Language Engineering, 27(1), 113–118.

Ding, B., Qin, C., Liu, L., Chia, Y. K., Li, B., Joty, S., & Bing, L. (2023, July). Is GPT3 a good data annotator? In A. Rogers, J. Boyd-Graber, & N. Okazaki (Eds.), Proceedings of the 61st annual meeting of the association for computational linguistics (volume 1: Long papers) (pp. 11173–11195). Association for Computational Linguistics. https://doi.org/10.18653/v1/2023.acl-long.626

Gagrčin, E., Porten-Cheé, P., Leißner, L., Emmer, M., & Jørring, L. (2022). What makes a good citizen online? the emergence of discursive citizenship norms in social media environments. Social Media+ Society, 8(1), 20563051221084297.

Gilardi, F., Alizadeh, M., & Kubli, M. (2023). Chatgpt outperforms crowd workers for text-annotation tasks. Proceedings of the National Academy of Sciences, 120(30), e2305016120.

Goyal, N., Kivlichan, I. D., Rosen, R., & Vasserman, L. (2022). Is your toxicity my toxicity? exploring the impact of rater identity on toxicity annotation. Proc. ACM Hum.-Comput. Interact., 6(CSCW2). https://doi.org/10.1145/3555088

Huang, F., Kwak, H., & An, J. (2023). Is chatgpt better than human annotators? potential and limitations of chatgpt in explaining implicit hate speech. Companion proceedings of the ACM web conference 2023, 294–297.

Kocoń, J., Cichecki, I., Kaszyca, O., Kochanek, M., Szydło, D., Baran, J., Bielaniewicz, J., Gruza, M., Janz, A., Kanclerz, K., et al. (2023). Chatgpt: Jack of all trades, master of none. Information Fusion, 99, 101861.

Krippendorff, K. (1970). Estimating the reliability, systematic error and random error of interval data. Educational and psychological measurement, 30(1), 61–70.

Larimore, S., Kennedy, I., Haskett, B., & Arseniev-Koehler, A. (2021, June). Reconsidering annotator disagreement about racist language: Noise or signal? In L.-W. Ku & C.-T. Li (Eds.), Proceedings of the ninth international workshop on natural language processing for social media (pp. 81–90). Association for Computational Linguistics. https://doi.org/10.18653/v1/2021.socialnlp-1.7

Long, C. (2021). How 'let's go brandon'became code for insulting joe biden. AP News, 30.

Marvin, G., Hellen, N., Jjingo, D., & Nakatumba-Nabende, J. (2023). Prompt engineering in large language models. International Conference on Data Intelligence and Cognitive Informatics, 387–402.

Mathew, B., Saha, P., Tharad, H., Rajgaria, S., Singhania, P., Maity, S. K., Goyal, P., & Mukherjee, A. (2019). Thou shalt not hate: Countering online hate speech. Proceedings of the international AAAI conference on web and social media, 13, 369–380.

Mellon, J., Bailey, J., Scott, R., Breckwoldt, J., Miori, M., & Schmedeman, P. (2024). Do ais know what the most important issue is? using language models to code open-text social survey responses at scale. Research & Politics, 11(1), 20531680241231468.

Mirza, A., Alampara, N., Kunchapu, S., Emoekabu, B., Krishnan, A., Wilhelmi, M., Okereke, M., Eberhardt, J., Elahi, A. M., Greiner, M., et al. (2024). Are large language models superhuman chemists? arXiv preprint arXiv:2404.01475.

Ollion, E., Shen, R., Macanovic, A., & Chatelain, A. (2023). Chatgpt for text annotation? mind the hype. Plank, B. (2022, December). The "problem" of human label variation: On ground truth in data, modeling and evaluation. In Y. Goldberg, Z. Kozareva, & Y. Zhang (Eds.), Proceedings of the 2022 conference on empirical methods in natural language processing (pp. 10671–10682). Association for Computational Linguistics. https://doi.org/10.18653/v1/2022.emnlp-main.731

Polak, M. P., & Morgan, D. (2024). Extracting accurate materials data from research papers with conversational language models and prompt engineering. Nature Communications, 15(1), 1569.

Reiss, M. V. (2023). Testing the reliability of chatgpt for text annotation and classification: A cautionary remark. arXiv preprint arXiv:2304.11085.

Rescala, P., Ribeiro, M. H., Hu, T., & West, R. (2024). Can language models recognize convincing arguments? arXiv preprint arXiv:2404.00750.

Rossini, P. (2022). Beyond incivility: Understanding patterns of uncivil and intolerant discourse in online political talk. Communication Research, 49(3), 399– 425.

Sap, M., Swayamdipta, S., Vianna, L., Zhou, X., Choi, Y., & Smith, N. A. (2022, July). Annotators with attitudes: How annotator beliefs and identities bias toxic language detection. In M. Carpuat, M.-C. de Marneffe, & I. V. Meza Ruiz (Eds.), Proceedings of the 2022 conference of the north american chapter of the association for computational linguistics: Human language technologies (pp. 5884–5906). Association for Computational Linguistics. https://doi. org/10.18653/v1/2022.naacl-main.431

Scheibenzuber, C., Neagu, L.-M., Ruseti, S., Artmann, B., Bartsch, C., Kubik, M., Dascalu, M., Trausan-Matu, S., & Nistor, N. (2023). Dialog in the echo chamber: Fake news framing predicts emotion, argumentation and dialogic social knowledge building in subsequent online discussions. Computers in Human Behavior, 140, 107587.

Schöpke-Gonzalez, A. M., Atreja, S., Shin, H. N., Ahmed, N., & Hemphill, L. (2022). Why do volunteer content moderators quit? burnout, conflict, and harmful behaviors. New Media & Society, 14614448221138529.

Schulman, J., Zoph, B., Kim, C., Hilton, J., Menick, J., Weng, J., Uribe, J. F. C., Fedus, L., Metz, L., Pokorny, M., et al. (2022). Chatgpt: Optimizing language models for dialogue. OpenAI blog, 2, 4.

Shea, A. L., Omapang, A. K. B., Cho, J. Y., Ginsparg, M. Y., Bazarova, N., Hui, W., Kizilcec, R. F., Tong, C., & Margolin, D. (2024). Discursive objection strategies in online comments: Developing a classification schema and validating its training.

Thapa, S., Naseem, U., & Nasim, M. (2023). From humans to machines: Can chatgptlike llms effectively replace human annotators in nlp tasks. Workshop Proceedings of the 17th International AAAI Conference on Web and Social Media.

Vahed, S., Goanta, C., Ortolani, P., & Sanfey, A. G. (2024). Moral judgment of objectionable online content: Reporting decisions and punishment preferences on social media. Plos one, 19(3), e0300960.

Vitak, J., Chadha, K., Steiner, L., & Ashktorab, Z. (2017). Identifying women's experiences with and strategies for mitigating negative effects of online harassment. Proceedings of the 2017 ACM Conference on Computer Supported Cooperative Work and Social Computing, 1231–1245.

Wan, Y., & Thompson, K. M. (2022). Making a cocoon: The social factors of pandemic misinformation evaluation. Proceedings of the Association for Information Science and Technology, 59(1), 824–826.

Wang, S., Liu, Y., Xu, Y., Zhu, C., & Zeng, M. (2021, November). Want to reduce labeling cost? GPT-3 can help. In M.-F. Moens, X. Huang, L. Specia, & S. W.-t. Yih (Eds.), Findings of the association for computational linguistics: Emnlp 2021 (pp. 4195–4205). Association for Computational Linguistics. https : / / doi . org/10.18653/v1/2021.findings-emnlp.354

Waterloo, S. F., Baumgartner, S. E., Peter, J., & Valkenburg, P. M. (2018). Norms of online expressions of emotion: Comparing facebook, twitter, instagram, and whatsapp. New media & society, 20(5), 1813–1831.

Wei, J., Wang, X., Schuurmans, D., Bosma, M., Xia, F., Chi, E., Le, Q. V., Zhou, D., et al. (2022). Chain-of-thought prompting elicits reasoning in large language models. Advances in neural information processing systems, 35, 24824–24837.

West, P., Lu, X., Dziri, N., Brahman, F., Li, L., Hwang, J. D., Jiang, L., Fisher, J., Ravichander, A., Chandu, K., et al. (2023). The generative ai paradox:"what it can create, it may not understand". The Twelfth International Conference on Learning Representations.

White, J., Fu, Q., Hays, S., Sandborn, M., Olea, C., Gilbert, H., Elnashar, A., SpencerSmith, J., & Schmidt, D. C. (2023). A prompt pattern catalog to enhance prompt engineering with chatgpt. arXiv preprint arXiv:2302.11382.

Wu, S., & Resnick, P. (2021). Cross-partisan discussions on youtube: Conservatives talk to liberals but liberals don't talk to conservatives. Proceedings of the International AAAI Conference on Web and Social Media, 15, 808–819.

Yin, Y., Jia, N., & Wakslak, C. J. (2024). Ai can help people feel heard, but an ai label diminishes this impact. Proceedings of the National Academy of Sciences, 121(14), e2319112121.